\pdfoutput=1
\documentclass[conference]{IEEEtran}

\author{Mykola Glybovets, Sergii Medvid\\
Faculty of Informatics\\
National University of Kyiv-Mohyla Academy, Ukraine\\
\texttt{\{glib, s.medvid\}@ukma.edu.ua}
}
\usepackage{amssymb}
\usepackage{cite}
\usepackage{hyperref}
\usepackage[linesnumbered,ruled,vlined]{algorithm2e}
\usepackage{amsmath}
\usepackage{booktabs}
\usepackage{float}
\usepackage{graphicx}
\usepackage{url}

\title{MorphoNAS: Embryogenic Neural Architecture Search Through Morphogen-Guided Development}

\begin{document}

\maketitle

\begin{abstract}
While biological neural networks develop from compact genomes using relatively simple rules, modern artificial neural architecture search methods mostly involve explicit and routine manual work. In this paper, we introduce MorphoNAS (Morphogenetic Neural Architecture Search), a system able to deterministically grow neural networks through morphogenetic self-organization inspired by the Free Energy Principle, reaction-diffusion systems, and gene regulatory networks. In MorphoNAS, simple genomes encode just morphogens dynamics and threshold-based rules of cellular development. Nevertheless, this leads to self-organization of a single progenitor cell into complex neural networks, while the entire process is built on local chemical interactions.

Our evolutionary experiments focused on two different domains: structural targeting, in which MorphoNAS system was able to find fully successful genomes able to generate predefined random graph configurations (8–31 nodes); and functional performance on the CartPole control task achieving low complexity 6-7 neuron solutions when target network size minimization evolutionary pressure was applied. The evolutionary process successfully balanced between quality of of the final solutions and neural architecture search effectiveness. Overall, our findings suggest that the proposed MorphoNAS method is able to grow complex specific neural architectures, using simple developmental rules, which suggests a feasible biological route to adaptive and efficient neural architecture search.
\end{abstract}

\section{Introduction}

Biological brains are not assembled region-by-region, it undergoes development stages using a \textbf{compact genomic blueprint}. Tony Zador's \textit{genomic bottleneck} hypothesis \cite{zador_critique_2019} implies, that the genome, likely, does not encode a \textbf{full diagram} of the brain, rather a \textbf{compressed cluster} of development instructions; mostly rules, constraints, and the conditions these rules must satisfy. This results in the evolution of efficient and flexible architectures that are far more adaptable than existing artificial systems.

Moreover, Zador suggests \cite{zador_critique_2019} that the general machine learning community has largely deviated from the original biological inspiration. Zador argues that young animal brains are capable of effective learning without enormous numbers of labeled examples used by both supervised and unsupervised machine learning nowadays, and points to another important implication of genomic bottleneck: it suggests a path toward artificial neural networks capable of rapid learning \cite{zador_critique_2019}.

The \textbf{Free Energy Principle (FEP)}, introduced by Karl Friston \cite{friston_learning_2003}, brings theoretical foundation, extending Zador's idea by arguing that all living systems, from cells to brains, must minimize variational free energy. This results in emergent self-organization into predictable, low-entropy states via \textbf{perception (inference)} and \textbf{action (development or adaptation)}. In morphogenesis \cite{turing_chemical_1990} context, this results in \textbf{embryonic structures emerging from local interactions}, driven by morphogens dynamics and \textbf{gene regulatory networks (GRNs)}.

By merging these principles into an evolutionary computational system, it becomes possible to \textbf{evolve neural architectures} not as explicit graphs, but as \textbf{embryogenically grown phenotypes}, ruled by a compact genome which encodes \textbf{morphogen-based rules} and is guided by \textbf{FEP-inspired self-organization}. This implies bottom-up approach to building cybernetic organism, a novel artificial neural network in this case, complementing traditional top-down approaches from neuroscience and cybernetics \cite{friston_knowing_2015}.

In this work, we introduce an evolutionary computational system, MorphoNAS, capable of generating neural networks through morphogenetic self-organization. The developmental model utilizes the Free Energy Principle, reaction-diffusion systems, and gene regulatory networks.

This approach was supported based on experiments we performed in two domains: the generation of neural networks with pre-defined topological properties (graph targeting), where all randomly generated configuration targets were achieved in 100\% of cases; and functional control task in CartPole environment, where minimal but capable neural network architectures were evolved under evolutionary pressure involving network size constraints.

We hypothesize that complex neural architectures capable of solving tasks may emerge from simple, biologically relevant, developmental rules. We provide examples of how networks grown using the MorphoNAS method can be applied to real functional optimization tasks.

These results suggest that the proposed evolutionary-developmental approach has potential as a generic implementation for automated neural architecture search, suitable for a wide range of applied domains.

\section{Key Mappings from Theory to Model}

Based on the theoretical foundations described above, let us now introduce a computational model that implements morphogenetic development using self-organizing local interactions to develop neural architectures. The key mappings from theory to the proposed model are the following:

\subsection{Free Energy Principle}

The behavior of the individual cells is driven by "surprise" or variational free energy minimization, corresponding to the fate decisions (division, differentiation, and axon growth) a cell must make in reaction to its local morphogenetic environment. The cells operate to achieve homeostasis or predictability in their morphogenetic locality, which is the source of the action(s) they take that are solely dependent on local morphogen concentrations with the decisions implemented as threshold-based transitions.

\subsection{Reaction-Diffusion Systems}

The spatial distribution and dynamics of morphogens is modeled with a reaction-diffusion process. A morphogen is released by a cell, which then diffuses throughout the developmental field, potentially inhibiting other morphogens. Together, this creates concentration gradients which provide positional information for the cells. The morphogen concentration gradients reflect Turing patterns and are responsible for symmetry breaking and spatial organization \cite{turing_chemical_1990}.

\subsection{Gene Regulatory Networks (GRNs)}

The genome $G$ represents the parameters that encode morphogen secretion rates, diffusion profiles, inhibition interactions, and cellular response thresholds. This is corresponding to the regulatory logic of GRNs which control how cells deal with morphogenetic signals and their responses.

\section{MorphoNAS Framework}

The MorphoNAS framework presented in this paper is based on the following four components: \textbf{progenitors}, which are similar to stem cells, can divide and become differentiated; \textbf{neurons}, which are differentiated cells capable of growing axons and forming synapse-type connections; \textbf{morphogens}, chemical fields that direct cell growth, behavior, and decision making via concentration fields; and \textbf{axon growth}, which acts in a chemotactic-type manner, having axons grow in the direction of increasing morphogen concentration, representing a simplified form of neuronal connectivity.

\subsection{Morphogenetic Development Model}
\label{MorphoNAS_definition}

MorphoNAS can be defined as a framework to simulate neural architectural formation via biologically inspired developmental processes. The structure is defined formally as
\[
\text{MorphoNAS} = (G, \Lambda, \mathcal{C}, \mathcal{M}, \mathcal{S}, \mathcal{D}, \mathcal{I}, T)
\]
where each of the components captures an integral aspect of the morphogenetic developmental process:

\medskip

\textbf{Developmental Field ($\boldsymbol{\Lambda}$)} is the spatial environment in which the neural growth occurs:

\[
\Lambda = \{(i, j) \mid 1 \leq i \leq L_x, 1 \leq j \leq L_y\}
\]

adopting a toroidal topology (connecting opposite edges), serving to model continuous spatial development.

\medskip

$\mathcal{C}$ denotes the \textbf{cell types}: $\mathcal{C} = \{\text{Progenitor}, \text{Neuron}\}$.  \textbf{Progenitor cells} are defined as undifferentiated cells that can divide and differentiate to form neurons.  \textbf{Neuron cells} are defined as differentiated cells that can grow axons, and make synapse-like connections.

\medskip

\textbf{Genome (G)} encodes the developmental program as a compact set of morphogenetic rules, which include: $G_{\text{dim}} = (L_x, L_y) \in \mathbb{N}^2$, describing the spatial dimensions of the developmental field $\Lambda$; $G_{\text{morph}}$, which defines the morphogen specifications -- each specification describing the secretion rates for the appropriate cell type, the diffusion properties of each morphogen (encoded as a tensor), and the parameters concerning cross-inhibition; $G_{\text{fates}}$ describes the threshold values that govern progenitor cell division, progenitor cell differentiation, and neuron axon growth; $G_{\text{axon}}$, describes the axon extension parameters to be taken -- driving the connection thresholds, and maximum lengths of axons; and $G_{\text{iter}} \in \mathbb{N}$, that represents development time -- in discrete time steps.

\medskip

\textbf{Morphogens ($\boldsymbol{\mathcal{M}}$)} comprise chemical signals that guide development: $\mathcal{M} = \{m_1, m_2, \ldots, m_k\}$, and diffuse spatially across the developmental field, providing positional information.

\medskip

\textbf{Cellular State ($\boldsymbol{\mathcal{S}}$)}: $\mathcal{S}: \Lambda \rightarrow \mathbb{R}^{|\mathcal{M}|} \times (\mathcal{C} \cup \{\emptyset\})$ defines local morphogen concentrations, and a resident cell (if any), for each location.

\medskip

\textbf{Diffusion Dynamics ($\boldsymbol{\mathcal{D}}$)}: $\mathcal{D}: \mathcal{M} \rightarrow \mathbb{R}^{(m(\mu) \times n(\mu))}$ defines how each morphogen $\mu \in \mathcal{M}$ diffuses through the developmental field, using diffusion tensors.

\medskip

\textbf{Inhibition Dynamics ($\boldsymbol{\mathcal{I}}$)}: $\mathcal{I}: \mathcal{M} \times \mathcal{M} \rightarrow \mathbb{R}$ defines the cross-inhibition of morphogen behaviour, which can lead to complex patterns through competitive interactions.

\medskip

\textbf{Developmental Dynamics ($\boldsymbol{T}$)}: $T: \mathcal{S}(\Lambda) \rightarrow \mathcal{S}(\Lambda)$ defines the temporal change of the system, implementation of the self-organizing growth process at each time-step.

\medskip

\begin{figure}[H]
\centering
\includegraphics[width=1\linewidth]{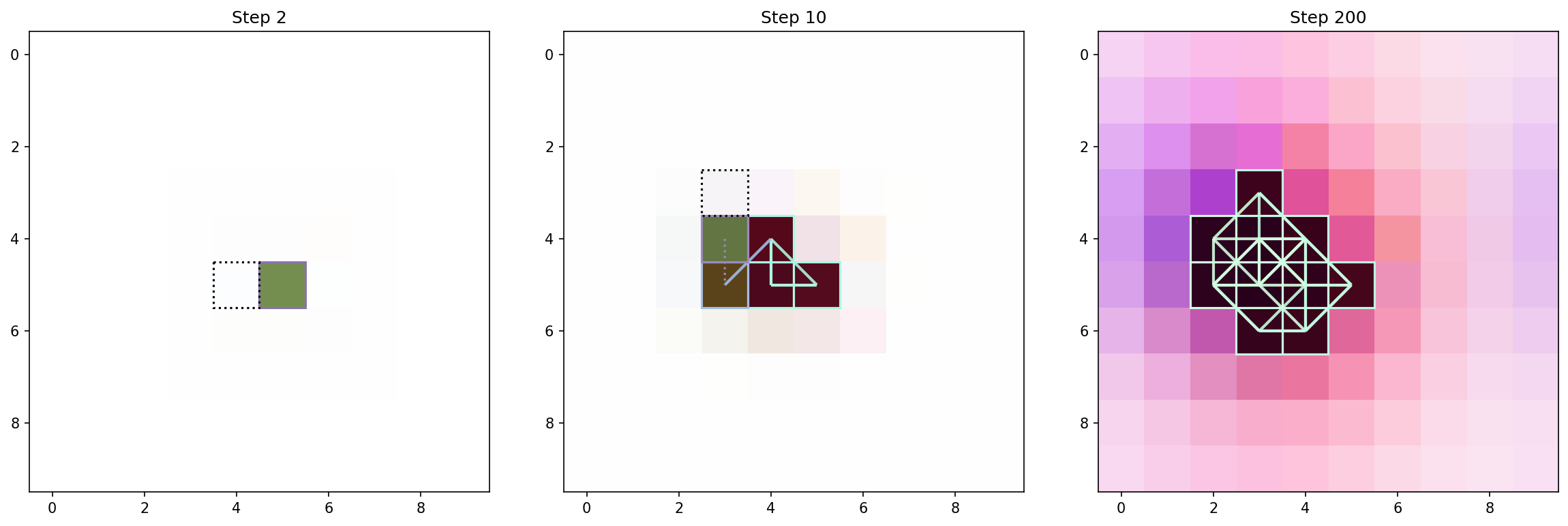}
\caption{Example of neural morphogenesis in the MorphoNAS system.}
\label{fig:MorphoNAS_example}
\end{figure}

An example of neural morphogenesis using the MorphoNAS system is shown in Figure~\ref{fig:MorphoNAS_example}. It demonstrates the self-organizing development of a neural network on a $10 \times 10$ developmental field after 2, 10 and, 200 simulated time steps resulting in 10 neurons and 36 connections. The gradients in the background of these simulations reflect the morphogen landscape achieved by the superposition of three chemicals.

\begin{figure}[H]
\centering
\includegraphics[width=1\linewidth]{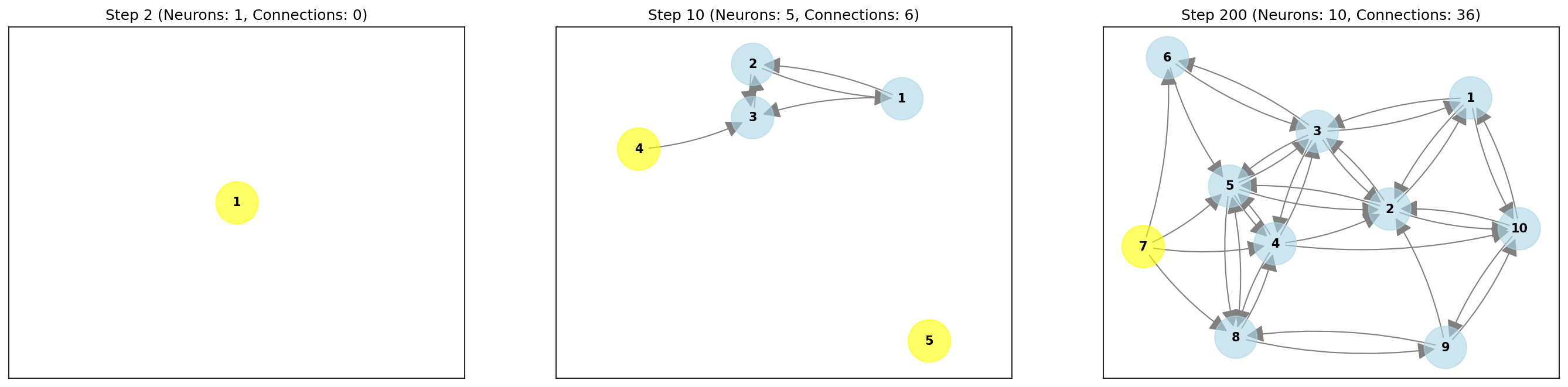}
\caption{Neural network structure at different stages of growth.}
\label{fig:MorphoNAS_example_graphs}
\end{figure}

The corresponding neural network structures at each stage of development are shown in Figure~\ref{fig:MorphoNAS_example_graphs}. At Step 2, the neural structure contains a single progenitor cell (not shown in the plot) and a single neuron. At Step 10, there are 4 neurons and 6 connections in total. There is also an isolated neuron disconnected from the cluster in the central region of growth field. By Step 200, there are 10 neurons and 36 connections developed.

\section{Morphogenetic Development Dynamics}

\subsection{Initialization}

At time $t = 0$, the developmental field $\Lambda$ is initialized with minimal structure: a single \textbf{progenitor} cell is placed at position $(i_0, j_0)$, like a neural stem cell; and all morphogen concentrations are set to zero in the entire developmental field.
\subsection{Iteration Steps}

At each iteration $t \in 0, 1, \ldots, G_{\text{iter}} - 1$, the following sequence is executed:
\begin{enumerate}
\item \textbf{Morphogen Secretion}: Cells release chemical signals into their local environment. For a cell of type $c$ at position $(i, j)$, morphogen concentrations are updated as: $c_m(i, j) \leftarrow c_m(i, j) + \sigma_c^{(m)}$, where $\sigma_c^{(m)}$ is the secretion rate of morphogen $m$ for cell type $c$.
    
    \item \textbf{Diffusion}: Morphogen concentrations are updated based on diffusion matrices according to one of several formulations typically using convolution or a numerical integration scheme (e.g., discrete Laplacian approximation): $c_m(i, j) \leftarrow \text{DiffusionStep}(c_m, \mathcal{D}(m))$
    
    \item \textbf{Inhibition}: Local interactions will modify concentrations according to the inhibition rules: $c_m(i, j) \leftarrow c_m(i, j) \prod_{n \neq m}(1 - \alpha_{mn} c_n(i, j))$, where $\alpha_{mn} = \mathcal{I}(m, n)$ is the inhibition coefficient of morphogen $n$ inhibiting morphogen $m$.
    
    \item \textbf{Cell Fate Decisions}: Each cell will evaluate its local morphogen environment in order to assess its developmental fate. When \textbf{progenitor cells} need to make decisions, they can exhibit three possible behaviors. They can (1) divide, in this case producing a new progenitor cell in an immediate neighbouring position; this happens if the concentration of morphogens that induce cell division exceeds a defined threshold; or (2) differentiate into a neuron, if the concentration of morphogens that are responsible for differentiation exceeds another predefined threshold; or (3) continue being in their current state if no cells or differentiation need to occur. \textbf{Neuron cells} will follow the morphogen gradients in the developmental environment to extend out axons towards neighbouring cells (other neurons) based on the chemical signals in their local environment.
    
    \item \textbf{Axon Growth and Connection}: Axons grow in a step-wise fashion, following the gradient ascent in morphogen fields. When a growing axon encounters another neuron it will form a stable connection if the local morphogen concentration exceeds the connection threshold $\theta_{\text{axon}} \in G_{\text{axon}}$.
    
    \item \textbf{Weight Assignment}: When the connection is formed, an initial synaptic weight \( w \) is assigned from morphogen concentration \( c \) at the target neuron and the distance \( d \) between the source neuron and target neuron. The weight calculation is:
\[
w = \max\left(0.01, \frac{c}{1 + d}\right)
\]
\end{enumerate}

The weights range between 0.01 and 1.0. The minimum weight of 0.01 ensures that the edge exists, while a weight of 0 indicates that no edge exists in the model.

Once edges are established, weights can change based on competitive scaling of the local concentrations of morphogens. Let \( c_{\text{local}} \) be the concentration of morphogen at the location of a neuron, while \( c_{\text{total}} \) is the total concentration of morphogen in the local neighborhood of this neuron. The new weight can be calculated as:
\[
w_{\text{new}} = \max\left(0.01, \frac{c_{\text{local}}}{c_{\text{total}}}\right)
\]

This creates a competitive situation where even relatively distant neurons with a high density of the axon-guiding morphogen will create more connections to other neurons than close neurons with a low density of morphogen.

\subsection{Termination}

This will continue for $G_{\text{iter}}$ iterations until the developmental process is terminated.

\medskip

Thus, the MorphoNAS system simulates the neural development over discrete time steps; where each location in the development field contains morphogen concentrations and possibly a cell (either progenitor or neuron). The system evolves step-wise and continues the procedure through secretion, diffusion, inhibition, cell fate decisions, and axon growth, according to local interactions and development rules encoded in the genome. Beginning with only one progenitor cell, the system self-organizes into a neural network through morphogen and chemical gradient signaling.  The complete process is detailed in Algorithm~\ref{alg:MorphoNAS}.
\begin{algorithm}
\label{alg:MorphoNAS}
\caption{Morphogenetic Development Dynamics}
\KwData{Initial developmental field $\Lambda$ with a single progenitor cell; genome $G$}
\KwResult{Final developmental state and neural network}
Initialize all morphogen concentrations to zero\;
\For{each time step $t = 1$ to $G_{\text{iter}}$}{
    \For{each cell $c$ in $\Lambda$}{
        Secrete morphogens at $c$ according to genome rules\;
    }
    \For{each morphogen $m$}{
        Diffuse $m$ across $\Lambda$ using diffusion matrix $\mathcal{D}(m)$\;
    }
    \For{each position $(i,j)$}{
        Apply inhibition between morphogens at $(i,j)$\;
    }
    \For{each cell $c$ in $\Lambda$}{
        \If{$c$ is a progenitor}{
            \uIf{division conditions are met}{
                Divide progenitor and place new progenitor in adjacent position\;
            }
            \uElseIf{differentiation conditions are met}{
                Differentiate progenitor into neuron\;
            }
        }
        \ElseIf{$c$ is a neuron}{
            Grow axon toward local morphogen gradient\;
            \If{axon reaches another neuron and connection threshold is satisfied}{
                Form connection and initialize synaptic weight\;
            }
        }
    }
}
\Return{final developmental state and neural network}
\end{algorithm}

\section{Evolutionary Targeting of Selected Graph Properties}

Having formally defined the MorphoNAS framework, parameterized with RD system characteristics and a set of simple GRN rules triggered solely by morphogen thresholds applied locally, let's assess its potential to produce graphs with predefined topological properties.

For this assessment a full isomorphic equality to a target graph is not feasible. First, despite recent developments, graph isomorphism is still a difficult computational problem \cite{babai_graph_2016}. Second, biological systems rarely evolve toward exact single structure \cite{edelman_degeneracy_2001}, rather exploring general properties that ensure functionality and robustness \cite{masel_robustness_2010}.

While in the proposed MorphoNAS one cell naturally self-organizes into complex graph-like structures, each genome deterministically defines a single graph. Evolutionary Algorithms (EAs) offer a powerful framework to explore the space of possible genomes, which will let us select for those that guide MorphoNAS process to form graphs that meet specific properties.

\subsection{Problem definition}

To evaluate the generative capacity of MorphoNAS without relying on exact graph matching (isomorphism), let's define a toy problem that specifies the following global properties of the resulting graph:

\begin{itemize}
    \item \textbf{Number of nodes}: The resulting graph should contain exactly $N_{\text{target}}$ nodes (neurons)
    \item \textbf{Number of edges}: It should contain exactly $E_{\text{target}}$ edges (synaptic connections)
    \item \textbf{Number of sources}: another easy-to-count property of the target graph is number of nodes with zero in-degree; the number of such nodes should match $S_{\text{target}}$
\end{itemize}

While the toy problem is limited to these 3 properties, other potential metrics could be explored in future research. Examples of such problems solvable in polynomial time compared to more complex graph isomorphism problem include mean shortest path length, graph density, presence of a giant component, and others.

\subsection{Fitness Function Design}

Now we can define a fitness function that will represent how closely the grown graph matches the target properties:
\begin{align*}
F(G) = &\ \exp\left(-w_N \cdot \frac{|N(G) - N_{\text{target}}|}{\text{tol}_N}\right) \\
&\times \exp\left(-w_E \cdot \frac{|E(G) - E_{\text{target}}|}{\text{tol}_E}\right) \\
&\times \exp\left(-w_S \cdot \frac{|S(G) - S_{\text{target}}|}{\text{tol}_S}\right) \\
&\times P(G)
\end{align*}

where $N(G), E(G), S(G)$ are the actual numbers of nodes, edges, and sources; $\text{tol}_N, \text{tol}_E, \text{tol}_S$ are tolerances that specify the degree of deviation; $w_N, w_E, w_S$ are weights indicating how much importance to place on each term.

$P(G)$ is a penalty for lack of weak connectivity defined as
\[
P(G) = \begin{cases}
1, & \text{if } G \text{ is weakly connected} \\
\gamma \leq 1, & \text{otherwise}
\end{cases}
\]

where $\gamma$ is a penalty factor, e.g. .1 or .5; if $\gamma$=1 then there is no penalty.

Some key features of the fitness function are: it returns a value of 1 when there is an exact match to target values; small deviations result in small reductions in fitness; as deviations increase, the penalties catch up exponentially. Finally, if the graph is not weakly connected, its fitness is multiplied by the factor $\gamma$ making it more difficult for disconnected graphs to survive

\subsection{Genetic Algorithm Setup}

\begin{algorithm}
\label{alg:ga}
\caption{Genetic Algorithm for Evolving MorphoNAS}
\KwData{Initial population of \( P \) genomes with fixed random seed}
\KwResult{Best evolved genome matching target graph properties}
Initialize population with random genomes\;
Evaluate fitness of all individuals in parallel\;
\While{generation $< G_{\mathrm{max}}$ \textbf{and} not converged}{
    Select \( P_{\text{parents}} \) parents using tournament selection (tournament size \( T_{\text{size}} \))\;
    Compute convergence ratio \( \rho = f_{\mathrm{avg}} / f_{\mathrm{best}} \)\;
    Update mutation multiplier \( \mu = \mu_{\mathrm{min}} + (\mu_{\mathrm{max}} - \mu_{\mathrm{min}}) \times \rho \)\;
    \For{\( P_{\text{replace}} \) offspring}{
        Randomly select two parents\;
        \eIf{parents have same morphogen count}{
            Inherit simple parameters randomly; perform element-wise crossover on secretion rates; perform row-wise crossover on inhibition matrix; inherit diffusion patterns from one parent\;
        }{
            Randomly choose one parent's morphogen count and associated parameters; randomly inherit other parameters\;
        }
        With probability \( p_{\mathrm{mut}} \times \mu \), apply mutation\;
        \If{mutation occurs}{
            Select mutation type according to probability vector \( \mathbf{v}_{\mathrm{mut}} \): grid size mutation (\( \pm 3 \)), growth steps mutation (\( \pm 20 \)), parameter mutation (90\% scale float by [0.5, 2.0], 10\% increment integer by \( \pm 2 \)), morphogen number mutation (\( \pm 2 \)), or matrix mutation (modify inhibition or diffusion patterns), 5\% chance of radical mutation
        }
    }
    Preserve \( P_{\text{elite}} \) elite individuals\;
    Replace \( P_{\text{replace}} \) individuals with new offspring\;
    Evaluate fitness of all new individuals (use fitness cache)\;
    Update best fitness and convergence statistics\;
}
\Return{Best evolved genome}
\end{algorithm}

For the experiment setup, a steady-state genetic algorithm \cite{de_jong_analysis_1975} is employed. 

\textbf{The initial population} consists of $P$ individuals, each having a genome set up as defined in chapter \ref{MorphoNAS_definition}. All initial individuals define developmental field of the same size $(L_x, L_y)$, the same number of morphogens $N_{\mathcal{M}}$, and the same maximum of $G_{\text{iter}}$ CA growth iterations. A fixed random seed ensured reproducibility of initial population.

\textbf{Parent selection} is being performed using tournament selection \cite{goldberg_comparative_1991} with a tournament size of $T_{\text{size}}$. The algorithm selection pressure is set to $p_{\text{sel}}$ for the entire run. As part of steady-state setup, $P_{\text{replace}}$ individuals get replaced in each generation. Elitism \cite[p. 101]{de_jong_analysis_1975} is used to preserve top $P_{\text{elite}}$ individuals without modification.

\textbf{Crossover} applies to the selected parents by randomly combining simple parameters from each genome. Morphogen secretion rates are mixed using element-wise crossover, inhibition matrices are recombined row-wise, and diffusion matrices are inherited in whole from one parent. When parents differ in morphogen count, the entire morphogen-specific parameters set is inherited by each offspring from one randomly selected parent, keeping other parameters mixed from both parents.

\textbf{Mutation} is applied with $p_{\text{mut}}$ probability to each produced offspring, selected randomly according to $\mathbf{v}_{\text{mut}}$ probabilities vector from five mutation types: developmental field size mutation, growth steps mutation, parameter mutation, morphogen number mutation, and matrix mutation. Parameter mutations target either floating-point parameters, scaling them by a random factor between 0.5 and 2.0 (90\% of cases), or integer parameters, changing them by $\pm 1$ (10\% of cases).

A "radical mutation" with 5\% chance is also introduced: when it occurs, the number of morphogens set to any in [3..13] range which results in all morphogen-related parameters randomized; the $G_{\text{iter}}$ growth steps set to random ones in [100..1000] range.

An \textbf{adaptive mutation strategy} was used for balancing exploration and exploitation. The effective mutation rate was scaled dynamically according to the population's convergence ratio, defined as the average fitness divided by the best fitness. The mutation rate multiplier $\mu$ was computed as follows:

\[
\mu = \mu_{\text{min}} + (\mu_{\text{max}} - \mu_{\text{min}}) \times \rho
\]

where $\mu_{\text{min}}$ and $\mu_{\text{max}}$ are the minimum and maximum scaling factors, respectively, and $\rho$ represents convergence ratio defined as the ratio of the average fitness $f_{\text{avg}}$ to the best fitness $f_{\text{best}}$ in the population: $\rho = \frac{f_{\text{avg}}}{f_{\text{best}}}$.

In each experiment run, the evolution process was running for up to $G_{\text{max}}$ generations or until convergence.

\subsection{Assessing the Suggested Evolutionary Framework}

To assess the suggested evolutionary framework, we will define $K$ target properties sets $\{N_{\text{target}}, E_{\text{target}}, S_{\text{target}}\}$ to define the desired number of neurons, edges, and sources per set of target properties.

The target properties are created by sampling of $K$ random directed graphs created with Erdős-Rényi $G(n, m)$ method \cite{erdos_random_1959} with a fixed $\text{seed}_{\text{target}}$. The $n$ and $m$ parameters are sampled randomly while ensuring there is a minimum average outdegree $\bar{k}_{\text{out}} \geq k_{\text{min}}$ and a maximum density of $d_{\text{max}}$.

For each target properties set, $R$ independent evolutionary runs were performed, each initialized with a different random seed, passed to Algorithm \ref{alg:ga} represented in pseudocode above. For the seeding process we create a sequence $s_{\text{runs}} = \{s_1, s_2, \ldots, s_R\}$, built with another fixed $\text{seed}_{\text{runs}}$ to allow each of the $K$ target properties set reuse this same sequence. Each of the evolutionary runs will be executing until a maximum number of generations $G_{\text{max}}$ is reached, or until the fitness threshold has been achieved, or until the average population fitness has converged to at least 95\% of the best fitness.

Performance of each run is defined by the best fitness recorded in that run, calculating the mean and standard deviation, while reporting the proportion of runs that achieve the fitness threshold.

Table~\ref{tab:full-parameters} contains the specific parameters for the experiment.

\begin{table}
\caption{Summary of Targeted Graph Properties Experiment Setup}
\label{tab:full-parameters}
\centering
\renewcommand{\arraystretch}{1.2}
\begin{tabular}{llp{5.5cm}}
\hline
\multicolumn{3}{c}{\textbf{Initial Genome Parameters}} \\
\hline
Symbol & Value & Description \\
\hline
\( (L_x, L_y) \) & (20, 20) & Developmental field dimensions (width × height) \\
\( N_{\mathcal{M}} \) & 3 & Number of morphogens \\
\( G_{\text{iter}} \) & 200 & Max growth iterations \\
\hline
\multicolumn{3}{c}{\textbf{Genetic Algorithm Parameters}} \\
\hline
\( P \) & 2000 & Population size \\
\( T_{\text{size}} \) & 7 & Tournament size \\
\( P_{\text{parents}} \) & 300 & Parents per generation \\
\( P_{\text{replace}} \) & 200 & Replaced individuals \\
\( P_{\text{elite}} \) & 10 & Elite individuals preserved \\
\( G_{\text{max}} \) & 1000 & Max number of generations \\
\( p_{\text{mut}} \) & 0.4 & Baseline mutation probability \\
\( \mu_{\text{min}} \) & 1.0 & Minimum mutation multiplier \\
\( \mu_{\text{max}} \) & 2.5 & Maximum mutation multiplier \\
\( \mathbf{v}_{\text{mut}} \) &
\shortstack[l]{[0.15, \\ 0.15,\\ 0.45, \\ 0.10,\\ 0.15]} &
Mutation type probabilities (grid size, steps, param, morphogens, matrix) \\
\( p_{rad} \) & 0.05 & Probability of radical mutation \\
\hline
\multicolumn{3}{c}{\textbf{Experimental Setup Parameters}} \\
\hline
\( K \) & 20 & Number of target property sets \\
\( R \) & 20 & Evolutionary runs per target \\
\( k_{\text{min}} \) & 1.5 & Min avg. out-degree \\
\( d_{\text{max}} \) & 0.3 & Max graph density \\
\( \text{seed}_{\text{target}} \) & 54210 & Seed for target graph generation \\
\( \text{seed}_{\text{runs}} \) & 65420 & Seed for run initialization \\
\hline
\end{tabular}
\end{table}

Table~\ref{tab:success_rates} shows the success rates of the evolutionary algorithm on 20 different target configurations, where each configuration was defined by a triplet $\{N_{\text{target}}, E_{\text{target}}, S_{\text{target}}\}$, defining the searched numbers of neurons, edges and source nodes. Success rate is the number of runs (out of $R$) that achieved a perfect fitness (the evolved graph specifically matched the properties that we required in our target configuration for the graph).

\begin{table}
\centering
\caption{Success rates across target sets sampled from random graphs.}
\label{tab:success_rates}
\begin{tabular}{lccc|c}
\toprule
\textbf{Target Set} & \(N_{\text{target}}\) & \(E_{\text{target}}\) & \(S_{\text{target}}\) & \textbf{Success Rate (\(R=20\))} \\
\midrule
K01 & 11 & 30 & 1 & 1.00 \\
K02 & 27 & 75 & 2 & 1.00 \\
K03 & 14 & 42 & 1 & 0.95 \\
K04 & 8 & 14 & 1 & 0.95 \\
K05 & 22 & 92 & 1 & 1.00 \\
K06 & 30 & 232 & 0 & 0.25 \\
K07 & 31 & 124 & 1 & 1.00 \\
K08 & 27 & 103 & 2 & 1.00 \\
K09 & 19 & 87 & 0 & 0.95 \\
K10 & 20 & 54 & 2 & 1.00 \\
K11 & 10 & 18 & 1 & 0.80 \\
K12 & 14 & 76 & 0 & 0.60 \\
K13 & 9 & 35 & 0 & 1.00 \\
K14 & 17 & 34 & 1 & 1.00 \\
K15 & 18 & 119 & 0 & 0.80 \\
K16 & 31 & 87 & 2 & 1.00 \\
K17 & 23 & 223 & 0 & 0.40 \\
K18 & 27 & 142 & 1 & 1.00 \\
K19 & 28 & 189 & 0 & 0.35 \\
K20 & 15 & 32 & 1 & 1.00 \\
\bottomrule
\end{tabular}
\end{table}

It is important to note that at least one valid solution was found for each target configuration, indicating the generality and robustness of our framework given a wide array of structural constraints. Several target sets (K06, K17, K19) did have lower success rates, but the algorithm still found a suite of valid successful solutions. This means that while the degree of search complexity has relatively increased, it has not failed the representational capability of genes, i.e. it has still been confirmed that these target configurations do have their respective genomes, despite they were found harder to locate for the EA. In addition, several configurations (K01, K05, K07) reliably produced 100\% of representations indicating the ease of convergence and the strength of the evolutionary algorithm stabilizing in those configurations.

\begin{figure}
\centering
\includegraphics[width=0.8\linewidth]{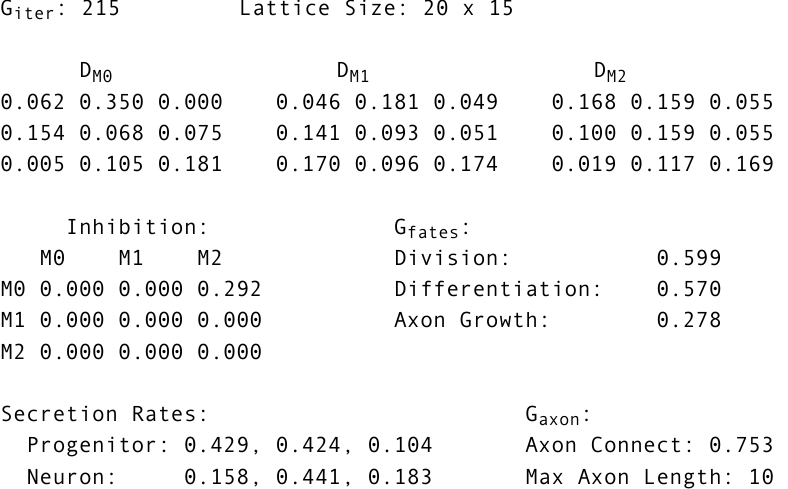}
\caption{Genome of a successful individual for target K16 (R07).}
\label{fig:genome_k16}
\end{figure}

In figure~\ref{fig:genome_k16}, we present the genome $G$ of a competent individual that was evolved to target K16 (run R07); K16 specifies a graph with the following properties: 31 neurons, 87 edges, and 2 inputs. The small parameter set represents growth dynamics, morphogen actions, cell fates and axon connectivity in a $20 \times 15$ developmental field.

This genome defines three morphogens, each with three distinct diffusion profiles: $(D_{M0}, D_{M1}, D_{M2})$, secretion rates for progenitor cells and neuron cells, and one non-zero cross-inhibition term. Cell behaviours are defined by thresholds in $G_{\text{fates}}$ for division, differentiation, and axon initiation. Axonal growth is managed by two parameters in $G_{\text{axon}}$: probability of connection, and maximum axon reach.

\begin{figure}
\centering
\includegraphics[width=0.5\linewidth]{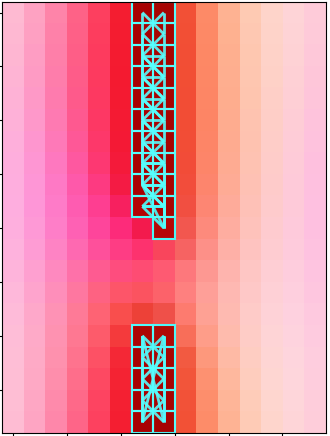}
\caption{Final state of the grown graph at step 215 for target K16 (R07), shown in a 2D projection of the toroidal developmental field topology.}
\label{fig:final_state_k16}
\end{figure}

In Figure~\ref{fig:final_state_k16}, we show the final grown graph state (step 215) in a 2D projection based on the toroidal topology of the developmental field.

For such a low dimensional structure, this genome has produced a valid offspring structure (neural graph) that has functional compliance (i.e. complex K16 specification), thereby illustrating the expressive power and generality of the evolutionary-developmental processes described here.

\begin{figure}
\centering
\includegraphics[width=0.7\linewidth]{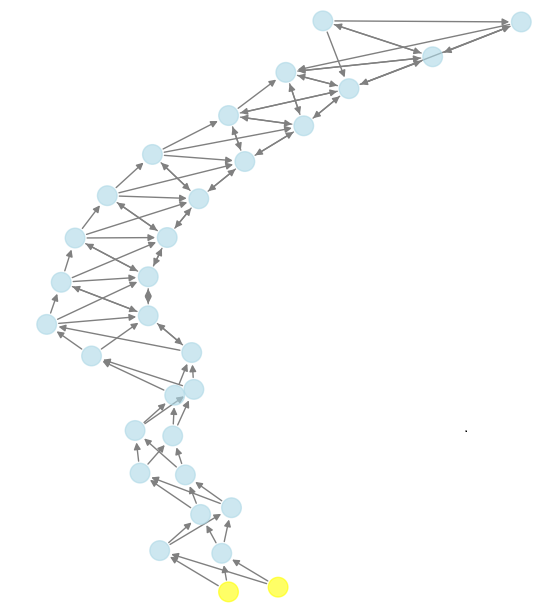}
\caption{Evolved graph for target K16 (R07)}
\label{fig:graph_k16}
\end{figure}

Figure~\ref{fig:graph_k16} provides the connectivity structure of the evolved graph that satisfies the K16 target specification.  Each node represents a neuron, each directed edge is associated with a network axon connection.  The two source nodes, shown at the bottom of the image serve as input neurons to the rest of the network and they have outgoing connectivity to other neurons in the network.  Overall, the topology is a complex but coherent wiring defined solely by the developmental process and the spatial and chemical constraints of the developmental field.  The graph is the direct result of the morphogen driven growth process as illustrated in the previous figure in the toroidal developmental field.

\subsection{Discussion}

The experiments showed that the proposed MorphoNAS Framework is capable of achieving its fundamental goal of defining simple RD- and GRN-only based genomes that yield neural networks with precise and specific structural targets. The fact that successful solutions were obtained across all 20 target configurations highlights not only the viability of the methods incorporated into this process, but also their generality for exploring different architectural constraints.

\textbf{Evolutionary Effectiveness:} The variability in success rates across targets (from more complex targets such as K06, K17, K19 compared to a flat out 100\% successful solutions like K01, or K05) is aligned with the inherent complexities of the target specifications, and is not a fundamental limitation of the framework. The fact that valid solutions could be found across all target configurations suggests that the proposed evolutionary-developmental framework has been able to explore the developmental programs effectively.

\textbf{Developmental Expressiveness:} The successful evolution on the example of target K16 (31 neurons, 87 edges, 2 source nodes) demonstrates the expressiveness of the proposed compact representation of a genome. The expressive power emerges from a low-dimensional parameter representation of the morphogen dynamics, cellular behaviours, and axonal growth rules, producing a reasonably complex network through self-organized developmental processes. This suggests that the proposed evolutionary-developmental principles, based on biological systems, can be useful generative models for neural architecture search.

\textbf{Scalability Indicators:} The ability of the framework to tackle varied target specifications which differed significantly in terms of not only scale, but also network connectivity density and structural complexity, suggest scalability is likely to be a promising area. Additionally, the spatial organization that emerges from morphogen-guided growth provides mechanism able to create structured connectivity patterns, which could be beneficial for more complex control tasks.

\textbf{Methodological Foundation:} These results provide evidence that the proposed process is capable of consistently converting high-level architectural specifications into working developmental programs. This capability forms a methodological foundation for functional validation that will be demonstrated in the next experiment described below, in which the evolved networks must both develop to achieve target structural properties and perform functional computations.

The broad success across target configurations confirms that the proposed evolutionary-developmental framework can be regarded as a viable mechanistic approach for automated neural architecture generation, and is now ready to tackle more advanced functional optimization problems.

\section{Applying Grown RNNs for Problem Solving}

Let's describe how the graphs grown using the proposed MorphoNAS framework can be applied to concrete problem-solving tasks.

After completing $G_{\text{iter}}$ iterations, the morphogenetic process yields a \textbf{directed weighted graph} $\mathcal{G} = (V, E, W)$, where: $V$ is the set of neurons (nodes); $E \subseteq V \times V$ is the set of directed edges (axon connections); and $W: E \rightarrow [0.01, 1.0]$ assigns a synaptic weight to each edge.

This graph defines a Recurrent Neural Network (RNN) structure on which it's required to deterministically define input and output neurons, applying propagation dynamics afterwards.

\subsection{Defining Input and Output Neurons}
\label{io_neurons}

First, let's define two sets of input and output neurons $V_{\text{in}} \subseteq V$ and $V_{\text{out}} \subseteq V$. The number of such neurons is determined by the dimensionality of the problem $d_{\text{in}}$ and $d_{\text{out}}$.

To make the neuron selection process deterministic, input neurons are assigned as the first \( d_{\text{in}} \) nodes of the grown graph sorted by nodes’ in-degree \( \deg^-(v) \):
\[
V_{\text{in}} = \{v_1, v_2, \ldots, v_{d_{\text{in}}}\} \subset V_{\text{sorted}},
\]

where \( V_{\text{sorted}} = \text{sort}(V, \text{by } \deg^-(v)) \).

Independently, the last \( d_{\text{out}} \) nodes are selected from the original grown graph order of $V$:
\[
V_{\text{out}} = \{v_{|V|-d_{\text{out}}+1}, \ldots, v_{|V|}\}.
\]

The following constraints are applied in the process:
\begin{itemize}
    \item $d_{\text{in}} + d_{\text{out}} \leq |V|$, so the emergent graph must be at least size of the problem's input plus output dimensions, and
    \item $V_{\text{in}} \cap V_{\text{out}} = \emptyset$, i.e. input and output neurons do not intersect.
\end{itemize}

\subsection{RNN Propagation Dynamics}
\label{propagation_dynamics}

Now, we will formalize the flow of information over the emergent directed graph. The graph is intended to be seen as a Recurrent Neural Network (RNN), in which neurons act as processing units and axon connections determine how information flows among neurons in the network. The propagation dynamics are modeled after \cite{najarro_towards_2023}. The flow of information progress through the network is discrete over a series of time steps and can be computed with matrix multiplications along with non-linear activation functions, over the graph structure.

Let

$\mathbf{x}_0 \in \mathbb{R}^{|V|}$ be the initial state of activation of the neurons, instantiated as follows:
\begin{itemize}
    \item Inputs are injected into the appropriate input neurons $V_{\text{in}} \subseteq V$, as defined above.
    \item All other neuron activations are initiated to the value of zero.
\end{itemize}

$\mathbf{W} \in \mathbb{R}^{|V| \times |V|}$ be the sparse weight matrix constructed from synaptic weights $W$ of the original graph:
\[
W_{ij} = \begin{cases}
W(i,j) & \text{if } (i,j) \in E \\
0 & \text{otherwise}
\end{cases}
\]

$f: \mathbb{R} \rightarrow \mathbb{R}$ be the neuron activation function (take, for example, $\tanh$), which would be computed for each element.

Let us now define the propagation of the network over $T$ discrete time steps (where $T$ is the diameter of $\mathcal{G}$, and optional additional timesteps are allowed):

For each time step $t \geq 0$:
\[
\mathbf{x}_{t+1} = \text{Update}(\mathbf{x}_t, f(\mathbf{W}\mathbf{x}_t))
\]

where $\text{Update}$ can operate in either of two modes: \textit{accumulation} mode implies $\mathbf{x}_{t+1} = \mathbf{x}_t + f(\mathbf{W}\mathbf{x}_t)$ or replacement mode implies $\mathbf{x}_{t+1} = f(\mathbf{W}\mathbf{x}_t)$.

\medskip

This mode of propagation has several characteristics. First of all, the process is completely deterministic and follows the structure of the grown graph. Also, an important feature is that the network can be run in a computationally efficient manner on both CPUs and GPUs, considering the sparsity of the dynamics. At the same time, the system supports both discrete and continuous input and output modalities. All this leads to the sustained state $\mathbf{x}_t$ changing in relation to the network topology and its weights, allowing for rich and expressive dynamic behavior.

\section{Evolving RNN Controllers for Gymnasium Environment}

\subsection{Problem Setting}

To confirm functional completeness, we will use the standard CartPole task from the Gymnasium environment \cite{towers_gymnasium_2024} as the most simple RL standard benchmark. In CartPole, the goal is to balance a vertical pole on a moving cart by applying a force to the left or right. The state of the system can be described by 4 numbers: cart position, cart speed, pole angle, and pole angular velocity. The agent can choose between just two discrete actions: apply left force and apply right force. The best (ideal) reward is 500 in this environment. 

The recurrent neural network (RNN) grown with MorphoNAS method will act as the cart controller. It accepts environtment state at its input, using the 4 input neurons $V_{in}$, and generates actions to stabilize the CartPole system using the 2 output neurons $V_{out}$.

\begin{figure}[H]
\centering
\includegraphics[width=0.5\linewidth]{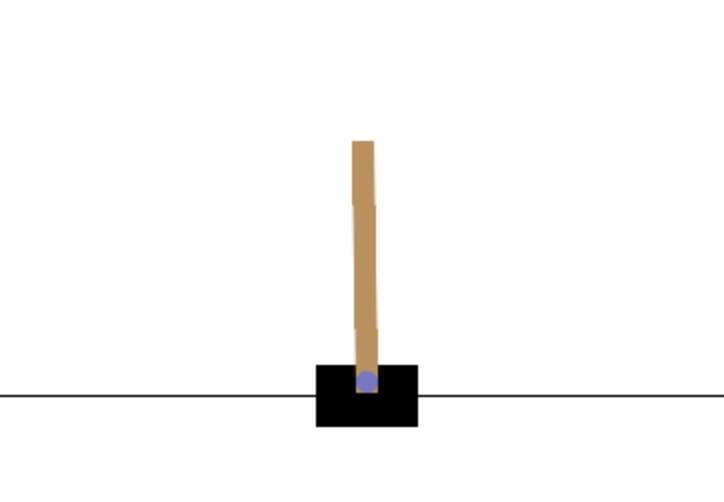}
\caption{CartPole from the Gymnasium environment}
\label{img:cartpole}
\end{figure}

\subsection{Fitness Function Design}
\label{fitness_function_design}

Let's define a fitness evaluation approach able to work with any standard environments from the Gymnasium library \cite{towers_gymnasium_2024}. The fitness function will determine how well a RNN grown by MorphoNAS performs on a set of random Gymnasium episodes. The performance a series of episodes is averaged and this score is normalized to give consistent fitness function outputs. 

The evaluation procedure can be executed on Gymnasium environments such as CartPole, LunarLander, MountainCar, and others. Each environment defines a predefined passing score which helps to perform normalization of the actual received score into a fitness function value.

The environment defines an observation space $\mathcal{O} \subseteq \mathbb{R}^{d_{\text{in}}}$ and an action space $\mathcal{A}$: either continuous $\mathbb{R}^{d_{\text{out}}}$ or discrete $\{0, 1, \ldots, k-1\}^{d_{\text{out}}}$.

The evaluation process will be defined for $N$ independent rollouts (episodes). For each rollout:

\begin{enumerate}
    \item \textbf{Input and output neurons are defined in the RNN}, as described in section \ref{io_neurons} above. The network may be so small it will trigger a fitness score of 0:
    \[
    F(\mathcal{G}) = 0, \text{ if } d_{\text{in}} + d_{\text{out}} > |V|
    \]
    
    \item \textbf{Inject observations.} At time step $t$, the environment provides observation $o_t \in \mathcal{O}$. The input neurons are initialized as $x_0[v_i] = o_t[i]$, for each $v_i \in V_{in}$ while all other neurons are set to zero.
    
    \item \textbf{Propagate.} The network propagates as defined in section \ref{propagation_dynamics}: $\mathbf{x}_{t+T} = \text{Propagate}(\mathbf{x}_0, \mathbf{W}, T)$.
    
    \item \textbf{Extract an action.} The action $a_t$ is extracted by the output neurons $a_t[i] = x_{t+T}[v_i]$, for each $v_i \in V_{\text{out}}$. Then depending on whether the action space is continuous or discrete, some postprocessing may be done: for discrete action spaces the actions are decided by $\arg\max$ or discretization, while for continuous action spaces the activations should be scaled to the valid action boundaries.
    
    \item \textbf{Step the environment.} Apply action $a_t$, receive reward $r_t$, and get the next observation $o_{t+1}$.
    
    \item \textbf{Accumulate rewards.} Finally, the rewards are accumulated across all steps in the episode:
    \[
    R^{(i)} = \sum_{t=0}^{T_{\text{episode}}^{(i)}} r_t^{(i)}
    \]
\end{enumerate}

The average reward across $N_{er}$ episode rollouts is:
\[
\bar{R} = \frac{1}{N_{er}} \sum_{i=1}^{N_{er}} R^{(i)}
\]

The average reward $\bar{R}$ is then converted to a fitness score $F(\mathcal{G}) \in [0,1]$ using a sigmoid transformation: 
\[
F(\mathcal{G}) = \frac{1}{1 + e^{-k(\bar{R} - S)}},
\]
where $S$ is the passing score for the given environment and $k > 0$ is the scaling factor that controls the slope of the sigmoid.

\subsection{Penalizing Excessive Networks}

Let's add an optional penalization to the fitness function so we can penalize excessive networks. It will cause the fitness to decrease down to the value of $\alpha$ (e.g. 0.8) as the number of connections increases. Let $N_C$ be the number of connections in the RNN, $\theta_C$ be the maximum number of unpenalized connections, and $\theta_{\text{half}}$ be the half-decay threshold (that is, the point where the fitness falls halfway between 1 and $\alpha$, i.e. to 0.9 if $\alpha = 0.8$). The decay rate is defined as $\lambda = \frac{\ln(2)}{\theta_{\text{half}} - \theta_C}$.

The connection penalization factor can then be defined as
\[
P_{\text{conn}}(N_C) = \alpha + (1 - \alpha) \times \exp(-\lambda \times \max(0, N_C - \theta_C)).
\]

And the penalized fitness function is then computed as
\[
F_{\text{eff}}(\mathcal{G}) = F(\mathcal{G}) \times P_{\text{conn}}(N_C).
\]

This discourages graphs with more than $\theta_C$ edges as it imposes a penalty on the fitness function which asymptotes back towards $\alpha$ as the number of connections increase.

\subsection{Genetic Algorithm Setup}

The setup for the evolution of RNN solving Gymnasium environments closely follows the parameterization described in Table~\ref{tab:full-parameters} for the Targeted Graph Properties experiment. The main difference lies in the fitness function, where the Gymnasium environment is employed as discussed in chapter \ref{fitness_function_design}. Other parameter modifications are provided in Table~\ref{tab:experiments-diff}.

\begin{table}
\centering
\caption{Experiments Parameters Difference}
\label{tab:experiments-diff}
\begin{tabular}{lccc}
\toprule
\textbf{Symbol} & \textbf{Targeting Graphs} & \textbf{RNN-Controller} & \shortstack{\textbf{RNN-Controller} \\ \textbf{(min)}} \\
\midrule
$F(\mathcal{G})$ & TargetGraphFitness & CartpoleFitness & CartpoleFitness \\
$(L_x, L_y)$ & (20, 20) & (10, 10) & (10, 10) \\
$P$ & 2000 & 500 & 500 \\
$P_{\text{parents}}$ & 300 & 100 & 100 \\
$P_{\text{replace}}$ & 200 & 50 & 50 \\
$p_{\text{mut}}$ & 0.4 & 0.3 & 0.3 \\
\midrule
\multicolumn{4}{c}{\textbf{Excessive Networks Penalization}} \\
\midrule
$\alpha$ & -- & -- & 0.8 \\
$\theta_C$ & -- & -- & 50 \\
$\theta_{\text{half}}$ & -- & -- & 1000 \\
\bottomrule
\end{tabular}
\end{table}

\subsection{Evaluation of Results}

For the first \textbf{RNN-Controller} experiment, a fixed seed, 65420, was used to generate $s_{\text{runs}}$ sequence of 100 seeds to randomize each run. An independent evolutionary run was launched for all seeds in $s_{\text{runs}}$.

It is worth to note that the ideal fitness for the CartpoleFitness function was found in the initial generation for 94 out of 100 runs; that is, no evolutionary steps were taken. The results are summarized in Table~\ref{tab:cartpole-generations}, outlining number of generations to reach optimal fitness in all runs.

\begin{table}
\centering
\caption{Number of runs reaching the ideal CartpoleFitness value with a population size of 500}
\label{tab:cartpole-generations}
\begin{tabular}{lc}
\toprule
\textbf{Time To Ideal Fitness} & \textbf{Number of Runs} \\
\midrule
Initial population (0 generations) & 94 \\
1 generation & 3 \\
4 generations & 1 \\
5 generations & 2 \\
\bottomrule
\end{tabular}
\end{table}

\begin{figure}
\centering
\includegraphics[width=0.8\linewidth]{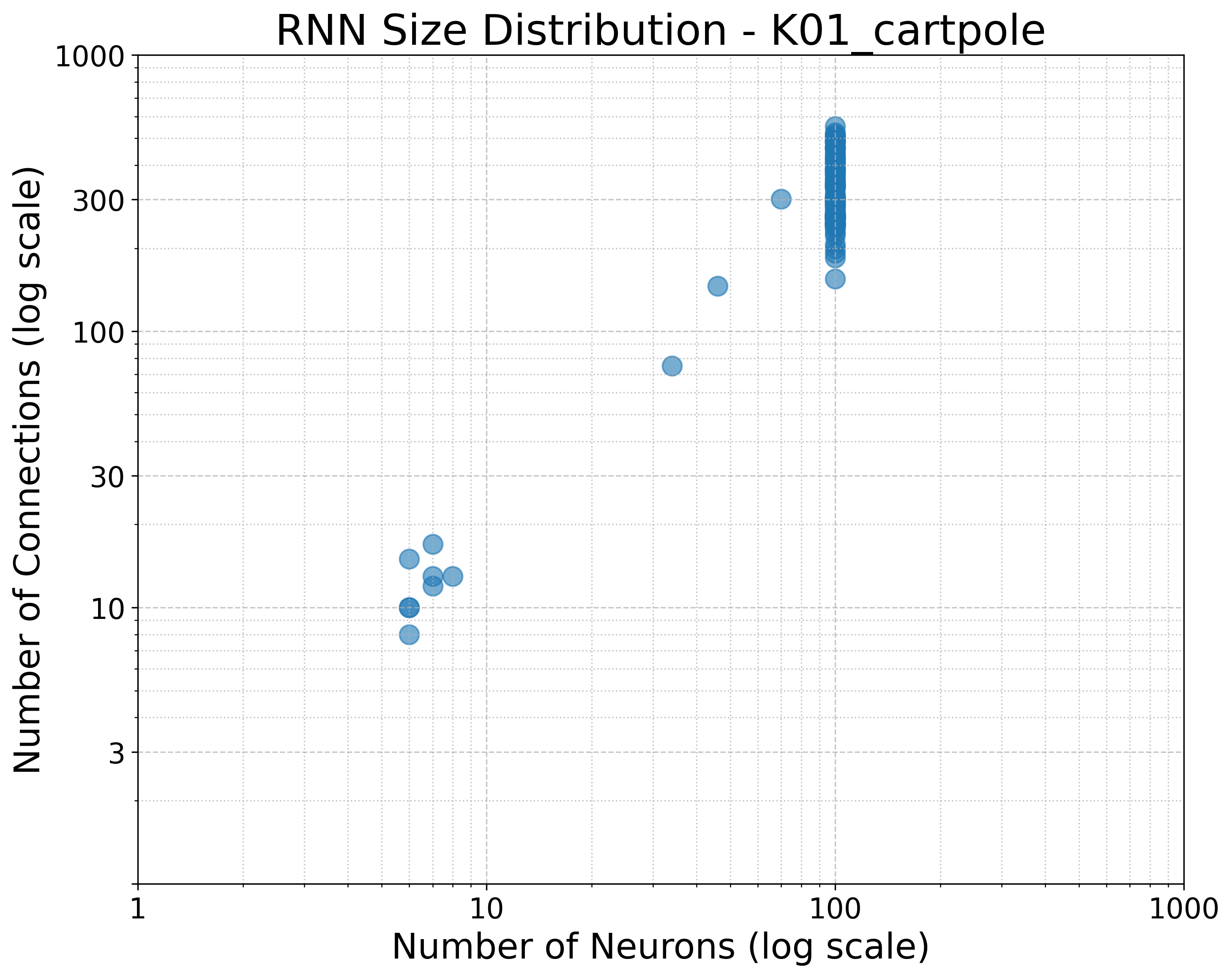} 
\caption{Sizes of RNN networks successfully operating CartPole found within Experiment B1}
\label{fig:rnn-size-dist-b1}
\end{figure}

In these 100 runs, the resulting RNNs contained fewer than 10 neurons in 8 cases, between 34 and 70 neurons in 3 cases, and exactly 100 neurons in the remaining 89 cases. It is shown on Figure~\ref{fig:rnn-size-dist-b1}.

For smaller search spaces, such as the Cartpole task from the Gymnasium suite, the MorphoNAS-based generation process typically yields RNNs achieving the ideal fitness with the maximum number of neurons, 100, for the developmental field of 10x10 size used in this experiment. For the Cartpole controller RNN, large networks (over 90 neurons) were observed in 89\% of populations.

In the \textbf{RNN-Controller (min)} version, we added a penalty into the fitness function to find smaller RNNs that could control CartPole with smaller numbers of nodes and connections. This shifted the distribution of found RNN CartPole controller networks with ideal fitness. The comparison is shown on the table. Intriguingly, 39\% of all runs in this experimental version yielded 6-neuron RNNs with ideal fitness when controlling CartPole while 33\% yielded 7-neuron RNNs.

\begin{table}
\centering
\caption{Summary Statistics of RNN CartPole Controller Experiments}
\label{tab:rnn-statistics}
\begin{tabular}{lcc}
\toprule
\textbf{Metric} & \textbf{RNN-Controller} & \shortstack{\textbf{RNN-Controller} \\ \textbf{(min)}} \\
\midrule
Small networks ($\leq 8$ neurons) & 8\% & 78\% \\
Medium networks (9-50 neurons) & 3\% & 20\% \\
Large networks ($\geq 90$ neurons) & 89\% & 2\% \\
\bottomrule
\end{tabular}
\end{table}

\begin{figure}
\centering
\includegraphics[width=0.8\linewidth]{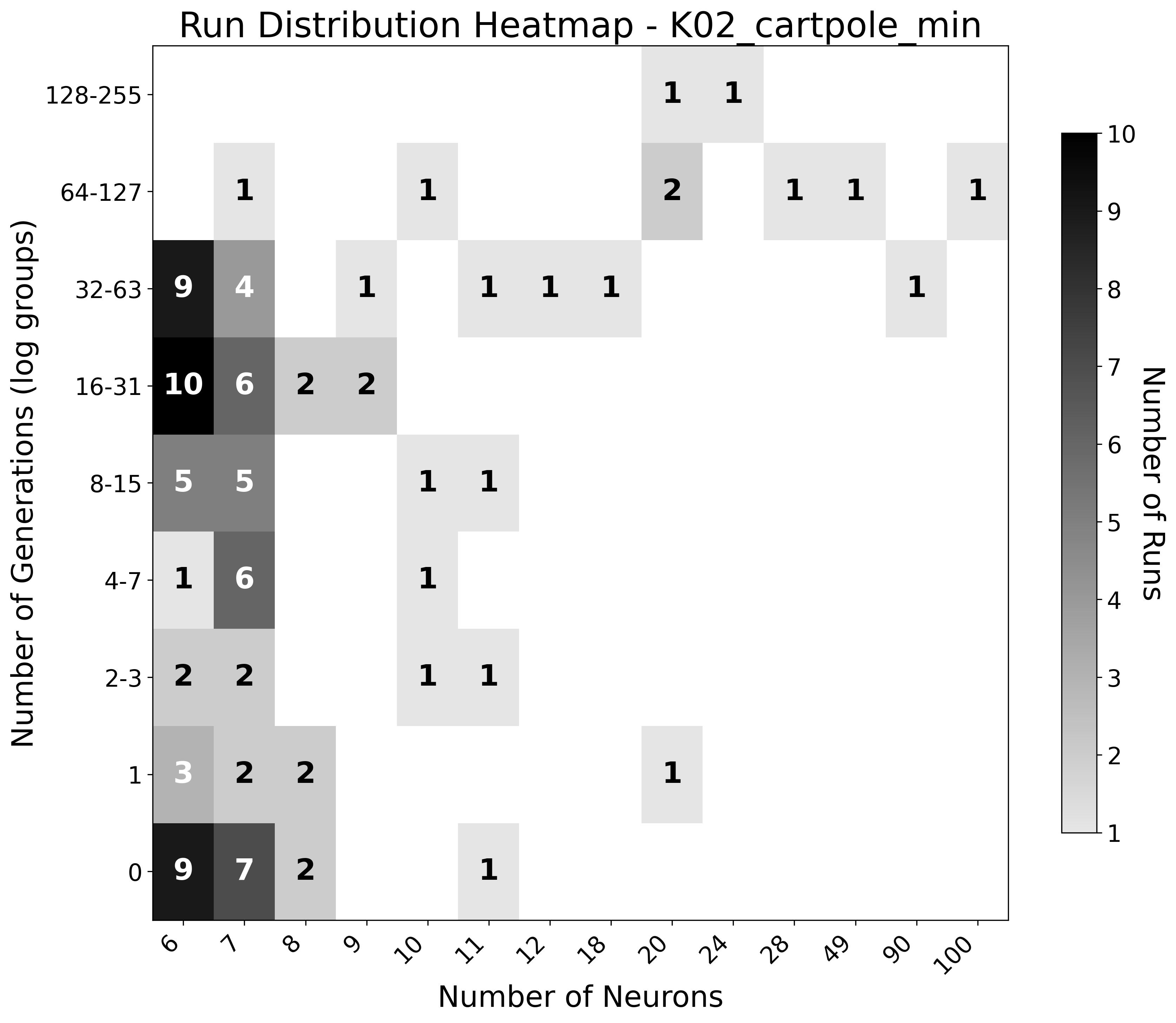} 
\caption{Distribution of successful evolutionary runs across network size and generation count. The values correspond to the number of runs resulting in finding a valid solution.}
\label{fig:heatmap-b2}
\end{figure}

The full distribution of RNN sizes for the \textbf{RNN-controller (min)} is shown in Figure~\ref{fig:heatmap-b2}. The values correspond to the number of evolutionary runs that had valid solutions. The heatmap shows that the compact solutions were not only occurring frequently, but also demonstrated consequtive patterns of evolutionary convergence, having highest concentration of successful solutions with 6 and 7 neurons. This clustering indicates that the network size-penalized fitness function provides enough evolutionary pressure to minimize the capacity of architecture while it is still able to accomplish the same function. It is interesting to note that most successful compact solutions required additional selection pressure to yield smaller neural networks.

\subsection{Discussion}

The results of \textbf{RNN Controller} experiment successfully demonstrate effectiveness of the proposed MorphoNAS method to produce functional control networks. High success rate, when 94 of 100 runs achieve optimal CartPole controller solustions in the initial population, suggests that the generation process effectively chooses viable network architectires.

It's important to point out that Oller et al. \cite{oller_analyzing_2020} have demonstrated that over 3\% of randomly weighted neural networks are able to operate CartPole successfully, meaning that CartPole is a relatively simple control task. However, this does not affect the  evaluation of the proposed experimental design: CartPole serves as a suitable proof-of-concept environment to test whether the MorphoNAS method can generate functional architectures, before we go up the complexity ladder. 

\textbf{Key Findings}: Our method demonstrates the ability to evolve different architectures depending on selective pressure. To underline, the initial \textbf{RNN Controller} experiment evolved functionally viable control networks that frequently had size of 100 neurons, maximum possible for the given developmental field size, while in the size penalised \textbf{RNN Controller (min)} experiment, 72\% of successful runs evolved a control network with 6-7 neurons, i.e., demonstrating that the size of the architecture was explicitly being considered by MorphoNAS. Correspondingly, there is a trade-off between solution quality and efficiency of architecture that the evolutionary process is clearly able to adapt to.

The emergence of compact, functional networks under selective evolutionary pressure indicates strong scalability prospects. The planned future experimental applications that involve more complex control tasks, including separate tasks for continuous physical control and partially observable environments, will help supply a rigorous experimental platform capable of providing sufficient assessment of the method's output capacity beyond this initial and descriptive verification.

The construction of both maximal and minimal viable architectures suggests that MorphoNAS-based processes can reliable grow functional neural networks within known solution spaces, thus laying the groundwork prior to entering and addressing much more complex problems.

\section{Conclusion}

In conclusion, this paper has established that MorphoNAS can produce neural networks with specified structural specifications and functional capabilities. The proposed biologically inspired alternative enhances standard neural architecture search by combining the Free Energy Principle, reaction-diffusion systems, and gene regulatory networks into one computational approach.

The experiment on \textbf{Graph Targeting}, in particular, demonstrated the generality of the system and produced valid solutions across random target configurations. Despite some of the configurations were more difficult to search than others, the successful solutions were found for all of them, showing potential for vaibility of the method across various architectural constraints.

The \textbf{RNN-controller} experiment illustrated functional competence of the system; where, at the population level, 94\% of the random populations contained networks that could successfully solve the CartPole task, demonstrating effective architecture sampling. At the same time, when network size penalties were added to the fitness function, evolution went on to find smaller and no less effective controllers where percentage of solutions having compact architecture with only 6-7 neurons was 72\%.

Integrating relevant principles from developmental biology and evolutionary computation lays the groundwork for automated generation of neural architectures. The key conclusion is that simple morphogen- and GRN-based rules can create complex neural architecture through self-organization.

In addition, this work demonstrates how morphogenetic systems can generate functional computational structures using a reduced set of biologically inspired developmental rules. It may also yield insights on the natural processes that occur in development of the neural tissue, and could potentially be beneficial to models in computational neuroscience.

While the validation was limited to simple control tasks with relatively small networks ($\leq 100$ neurons), the principles of spatial organization in the proposed framework promise scalability potential. Future work would engage computational optimization, expand the researched tasks to include more complex problem domains, and investigate hierarchical developmental processes for larger target architectures.

By incorporating plasticity mechanisms at development time or after development, MorphoNAS may be capable of producing networks that could then undergo further adaptation, potentially yielding a closer resemblance to biological neural systems. Also, a systematic study that compares MorphoNAS to the established neural architecture search methods would also be beneficial to clarify strengths, weaknesses, and appropriate domains of application.

As AI systems become more complex, bio-inspired developmental approaches may play significant role to discover neural architectures that maintain effectiveness, adaptability, and robustness of biological neural systems.

This preliminary validation marks MorphoNAS as an exciting approach in neural architecture search toolbox, and restates a principled pathway in pursuit of creating more biologically plausible artificial neural networks.

\section{Code and Data Availability}
The source code for MorphoNAS, including all experimental setups and evolved genomes, will be made publicly available at \url{https://github.com/sergemedvid/MorphoNAS} following publication. The repository will include: complete MorphoNAS framework implementation, evolutionary algorithm code, experimental configurations and seeds for reproducibility, example evolved genomes and visualizations.

\bibliographystyle{unsrt}
\bibliography{references}

\end{document}